\documentclass[10pt, a4paper]{article}

\usepackage{lipsum}
\usepackage{xcolor}
\usepackage{graphicx}
\usepackage{adjustbox}
\usepackage{tabularx}

\usepackage{booktabs}
\usepackage[final]{lrec2026} 

\title{Towards Corpus-Grounded Agentic LLMs for Multilingual Grammatical Analysis}

\name{
\begin{tabular}{c}
Matej Klemen$^1$, Tjaša Arčon$^1$, Luka Terčon$^2$, Marko Robnik-Šikonja$^1$, \\Kaja Dobrovoljc$^{1,2,3}$
\end{tabular}
} 

\address{$^{1}$University of Ljubljana, Faculty of Computer and Information Science, \\Večna pot 113, 1000 Ljubljana, Slovenia\\$^{2}$University of Ljubljana, Faculty of Arts,  Aškerčeva cesta 2, 1000 Ljubljana, Slovenia\\$^{3}$Jožef Stefan Institute, Jamova cesta 39, 1000 Ljubljana, Slovenia\\
         \{matej.klemen, tjasa.arcon, marko.robnik\}@fri.uni-lj.si, \\
         \{luka.tercon, kaja.dobrovoljc\}@ff.uni-lj.si}

\abstract{
Empirical grammar research has become increasingly data-driven, but the systematic analysis of annotated corpora still requires substantial methodological and technical effort. We explore how agentic large language models (LLMs) can streamline this process by reasoning over annotated corpora and producing interpretable, data-grounded answers to linguistic questions. We introduce an agentic framework for corpus-grounded grammatical analysis that integrates concepts such as natural-language task interpretation, code generation, and data-driven reasoning. As a proof of concept, we apply it to Universal Dependencies (UD) corpora, testing it on multilingual grammatical tasks inspired by the World Atlas of Language Structures (WALS). The evaluation spans 13 word-order features and over 170 languages, assessing system performance across three complementary dimensions -- dominant-order accuracy, order-coverage completeness, and distributional fidelity -- which reflect how well the system generalizes, identifies, and quantifies word-order variations. The results demonstrate the feasibility of combining LLM reasoning with structured linguistic data, offering a first step toward interpretable, scalable automation of corpus-based grammatical inquiry.
 \\ \newline  }

\begin{document}

\maketitleabstract

\section{Introduction}\label{sec:introduction}

\begin{figure*}[htbp]
    \centering
    \includegraphics[width=0.9\linewidth]{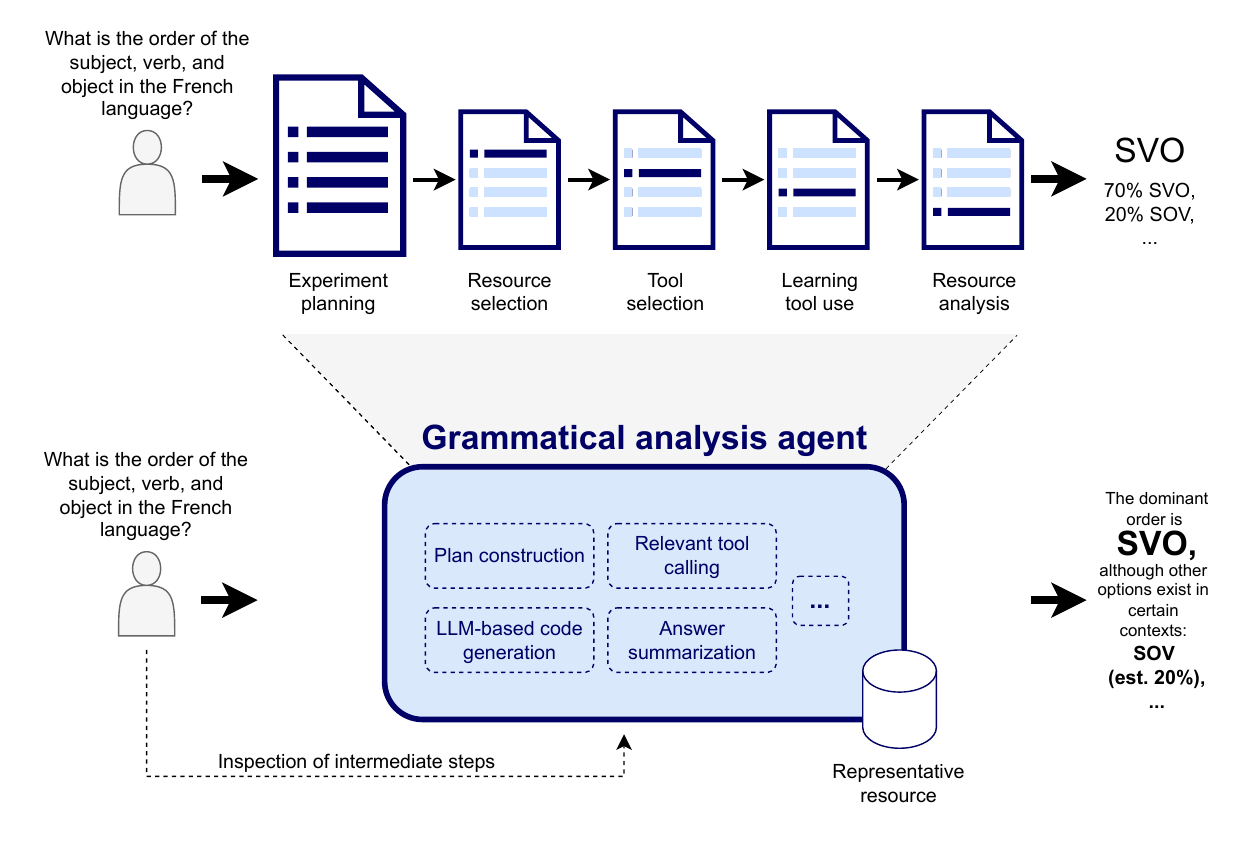}
    \caption{A simplified representation of the contrast between a traditional approach to corpus-based linguistics research (top), and our abstract solution proposal (Grammatical analysis agent, bottom).
    }
    \label{fig:motivation}
\end{figure*}

In the past four decades, linguistics has undergone a fundamental transformation from intuition-based inquiry to data-driven analysis, enabled by large-scale annotated corpora and computational tools that allow grammatical patterns to be studied systematically across languages and communicative contexts \cite{leech1992corpora,McEnery_Hardie_2011,Sinclair1991CorpusCC}. Yet, despite these advances, corpus-based linguistic analysis remains labor-intensive and hypothesis-driven, requiring complex scripting, specialized tools, readily accessible corpora, and methodological decisions that constrain exploratory research \cite{Gries2009,10.1093/oxfordhb/9780199544004.013.0008} -- limiting the scope of feasible investigations \cite{kaunisto2024challenges}.


 Recent advances in large language models (LLMs) with advanced reasoning capabilities \cite{Srivastava2023BeyondTI,10.5555/3600270.3602070} open new avenues for addressing these limitations, as they have been shown to capture a broad range of grammatical regularities, perform complex linguistic reasoning, and generate structured, interpretable outputs \cite{Begus_2025,jumelet2025multiblimp10massivelymultilingual,xia-etal-2024-fofo}. Their ability to understand natural-language instructions and interface with external data sources \citep{rag} suggests that they can now serve as intelligent mediators between linguistic questions and empirical evidence, bringing a new level of automation and accessibility to corpus-based grammatical analysis.

 

To illustrate the concept, we use the problem of determining the typical order of subject, verb, and object in a language throughout the paper. In Figure \ref{fig:motivation}, we show the contrast between the steps a user might take when trying to tackle the problem traditionally versus using an LLM-based grammatical analysis agent. In the traditional approach, the user needs to plan the experiment, select a relevant resource (e.g., a parsed corpus), select an appropriate tool (e.g., Python) and learn to use it, then analyze the resource. When using a grammatical analysis agent, the steps are abstracted away from the user, but may be inspected to validate the correctness of the process.

To explore this potential, we describe the evaluation framework to mimic the motivational use-case, and propose a grammatical agent prototype to test the feasibility of the idea. Specifically, we apply it to word-order analysis in the Universal Dependencies (UD) corpora, which provide cross-linguistically consistent grammatical annotation and have become a cornerstone for large-scale, data-driven studies of grammatical structure and variation \cite{10.1162/coli_a_00402,UD-2.16}. More concretely, our contributions are:
\begin{enumerate}
    \item \textbf{An exploratory evaluation framework for corpus-grounded grammar analysis.} We define a structured setup for evaluating grammatical analysis agents on word-order phenomena in UD data. The current demonstration framework covers 13 tasks across up to 179 languages, each with defined expected outputs. To enable automatic evaluation, we use classification accuracy, $F_1$ score, and Hellinger distance to measure different aspects of distributional grammatical analysis (dominant pattern prediction, coverage of relevant patterns, and pattern distribution prediction).
    
    \item \textbf{Baseline evaluation.} Using our framework, we test a strong baseline model (GPT-5) that approaches the tasks without explicit resource-grounding, instead relying on its pre-existing linguistic knowledge. We show that the model contains a varying amount of linguistic knowledge depending on the problem, but in general leaves significant room for improvement. 
    
    \item \textbf{Grammatical analysis agent prototype.} To verify the potential of resource grounding, we develop a prototype grammatical analysis agent (UDagent) with a simple agentic architecture: a fixed plan, a GPT-5-based code generation task-solving module, and answer construction based on the UD data. We show that despite being a conceptually simple prototype, the system already shows a performance improvement on most tasks.
\end{enumerate}


The remainder of this paper is organized as follows. Section \ref{sec:related-work} reviews related work. Section \ref{sec:task-and-data} introduces the task and evaluation framework including the design of the evaluation dataset. Section \ref{sec:system} describes the architecture of our agentic LLM system. Section \ref{sec:evaluation} presents evaluation results across models, features, and languages.
Section \ref{sec:conclusion} concludes with future directions.

\section{Related Work}\label{sec:related-work}

Corpus-based approaches have transformed grammatical research by enabling systematic analyses of authentic language data to uncover patterns, variation, and change \cite{biber2021grammar}. The growing availability of grammatically annotated corpora, such as those developed within the Universal Dependencies framework \cite{10.1162/coli_a_00402}, has been instrumental in advancing research across linguistic typology \cite{Levshina+2019+533+572,YanLiu2023,gerdes:hal-04067297}, data-driven grammar induction \cite{herrera-etal-2024-sparse}, cross-linguistic comparison of structural patterns \cite{doi:10.1073/pnas.1502134112}, and the study of syntactic complexity \citep{osti_10119528}. These developments have established a range of computational methods that require specialized technical and statistical expertise to extract and analyze grammatical patterns from parsed corpora.

More recently, advances in large language models have opened new possibilities for interacting with linguistic corpora through natural language. Initial applications demonstrate this potential in different contexts: the integration of LLMs into English-Corpora.org \cite{davies2025aillms} enables users to send corpus outputs such as frequency lists or concordance lines to an LLM for summarization, explanation, and interpretation; CorpusChat supports interactive language learning through corpus-informed dialogue grounded in authentic usage examples \cite{Cheung20052025}, highlighting the complementarity between corpus-based and generative AI approaches \cite{CROSTHWAITE2023100066}; and text-to-CQL systems translate natural-language requests into formal corpus queries \cite{lu2024textcqlbridgingnatural}. Beyond querying, LLMs have also shown promise in explaining linguistic patterns \cite{singh-etal-2023-explaining} and inducing grammatical rules from corpus data \cite{spencer-kongborrirak-2025-llms}. However, most of these applications remain task-specific and limited in scope, lacking a general, systematically evaluated framework that combines natural-language prompting with corpus-grounded reasoning to support large-scale grammatical analysis.

More broadly, the recent rise of agentic LLMs has transformed how language models are used for research, enabling them to autonomously plan tasks, access external data, and perform complex analyses across disciplines. The ReAct paradigm \citep{yao2023react} introduced an explicit coupling of reasoning and acting, allowing LLMs to interleave natural language reasoning steps with concrete tool- or environment-driven actions. Further work has focused on refining and expanding the agentic components, often focusing on a particular use case such as data analysis \citep{ds-agent} or machine learning \citep{ml-agent}. In our work, we explore grammatical analysis as a new application for LLM-based agents, present a prototype system UDagent, and an evaluation framework along with a multilingual dataset to support further development.



\section{Task and Evaluation Framework}\label{sec:task-and-data}

\begin{table*}[htbp]
\centering
\begin{tabular}{lp{6.5cm}ccc}
\toprule
\textbf{WALS ID} & \textbf{Linguistic Task} & \textbf{\#Options} & \textbf{\#Languages} & \textbf{\#Languages$_{>30}$}\\      
\midrule
81A & Order of subject, object and verb & $6$ & $169$ & $92$ \\
82A & Order of subject and verb & $2$ & $177$ & $147$ \\
83A & Order of object and verb & $2$ & $178$ & $156$ \\
84A & Order of object, oblique, and verb & $6$ & $165$ & $101$ \\
85A & Order of adposition and noun phrase & $2$ & $172$ & $133$ \\
86A & Order of genitive and noun & $2$ & $107$ & $66$ \\
87A & Order of adjective and noun & $2$ & $163$ & $123$ \\
88A & Order of demonstrative and noun & $2$ & $104$ & $74$ \\
89A & Order of numeral and noun & $2$ & $163$ & $101$ \\
90A & Order of relative clause and noun & $2$ & $158$ & $104$ \\
94A & Order of adverbial subordinator and clause & $3$ & $160$ & $93$ \\
144A & Order of subject, object, verb, and negative word & $25$ & $76$ & $18$* \\
144B & Order of negative word and verb relative to clause boundaries & $6$ & $94$ & $53$ \\
\bottomrule
\end{tabular}
\caption{Linguistic questions in the evaluation dataset along with the number of possible answer options (\#Options), number of languages for which at least one valid answer was produced using the ground truth scripts (\#Languages), and number of those for which at least $30$ answers were produced (\#Languages$_{>30}$). 
}
\label{tab:all-ling-questions-compressed}
\end{table*}


Building on the conceptual framework presented in Figure \ref{fig:motivation}, this section describes how the approach was instantiated and evaluated on word-order phenomena in the Universal Dependencies (UD) corpora. We first define the grammatical task at hand (Section \ref{sec:task-definition}), describe the evaluation dataset and its cross-linguistic coverage (Section \ref{sec:evaluation-method}), and then introduce the evaluation metrics used to assess system performance (Section \ref{sec:evaluation-metrics}).

\subsection{Task Definition}\label{sec:task-definition}

The evaluation framework builds on typological distinctions captured in the World Atlas of Language Structures (WALS; \citetlanguageresource{wals}) which documents cross-linguistic variation across hundreds of grammatical features based on evidence from descriptive grammars. WALS encodes how individual languages realize specific linguistic properties, assigning each language one of several predefined categorical values. For example, for the feature \textit{Order of Object and Verb}, each language is classified as \textit{VO} (the object follows the verb), \textit{OV} (the object precedes the verb), or\textit{ No dominant order}.

We take this typological framework as a starting point but extend it to reflect patterns actually attested in corpus data. While WALS adopts discrete, classification-oriented labels, many languages exhibit within-language variation that is better captured through gradient, corpus-based evidence \cite{YanLiu2023,Levshina2023,Baylor2024}. Our formulation therefore complements WALS-like grammatical description by producing richer, multi-layered outputs that reveal not only the dominant order but also the diversity and proportional distribution of patterns observed in real usage.

Each grammatical analysis task corresponds to analysing one such word-order relation by receiving as input a grammatical question and a target language, such as: “What is the order of subject, object, and verb in French?” Given such a question and the corresponding language corpus, the system is expected to produce a structured summary of the observed patterns. The output consists of three layers of analysis, listed from least to most detailed:

\begin{itemize}
    \item  \textbf{Dominant order:} the most frequent pattern in the corpus (e.g., SVO);
    \item  \textbf{Attested orders:} all observed patterns in the corpus (e.g., SVO, SOV, VSO);
    \item \textbf{Order distribution:} proportional frequencies of each pattern in the corpus (e.g., SVO: 92\%, SOV: 6\%, VSO: 2\%)
\end{itemize}

This three-layered formulation defines grammatical analysis as a structured, data-driven reasoning task that enables evaluation across increasing levels of descriptive completeness. Although developed to support the word-order analyses presented in the continuation of this paper, in further work, the same reasoning framework will be extended to other tasks that examine the distribution and variation of linguistic patterns in corpus data.

\subsection{Evaluation Dataset}\label{sec:evaluation-method} 


Building on the word-order analysis framework introduced in Section \ref{sec:task-definition}, we constructed a large-scale multilingual dataset comprising 13 word-order-related grammatical questions with three-layer ground-truth annotations derived from the UD corpora. UD is a collection of manually annotated treebanks that provide consistent morphosyntactic annotation across typologically diverse languages \citep{10.1162/coli_a_00402}, with each word analyzed for lemma, part of speech, morphological features, and syntactic head–dependent relations following a uniform dependency annotation scheme.



\textbf{Task selection.} 
To align our work with aforementioned typological investigations, we reviewed all 56 Word Order features documented in WALS, and selected those meeting three criteria: (1) fully representable using UD annotation, (2) not dependent on language-specific lexical information, (3) compatible with our three-step analytical framework. 
This yielded a total of 13 features, each reformulated as a natural language question for evaluation (e.g., \textit{What is the order of subject, object, and verb in language X?}).
Possible answer categories were based on WALS inventories, with minor adaptations to ensure compatibility with UD data and to maintain a consistent analytical framework (e.g., the “No dominant order” category was omitted in favor of a simplified approach that always selects the most frequent order, as described below).

 
\textbf{Ground-truth generation.} Gold-standard answers were derived automatically from test sets of 179 UD (v2.16; \citetlanguageresource{11234/1-5901}) treebanks (one per language) using a human-written Python script. When selecting a treebank for a language, we prioritize larger treebanks that span diverse domains\footnote{We will provide the specific treebanks used in the Github repository after the review period.}. 
For each task, the scripts counted all configurations corresponding to a given question and produced two complementary outputs: (1) a list of attested order variations with their frequencies, and (2) a final answer, determined by selecting the most frequent order variation. 



\textbf{Dataset composition.} The final evaluation dataset consists of the 13 linguistic questions, their predefined value sets, and the corresponding three-layered gold answers across 175 languages, resulting in 2,275 individual feature-language pairs for benchmarking future grammatical analysis systems. The dataset is openly released to support reproducible experimentation and cross-linguistic comparison.\footnote{Anonymized for review.}


\textbf{Evaluation scope.} Table \ref{tab:all-ling-questions-compressed} summarizes all linguistic tasks (questions) included in the evaluation dataset, showing their corresponding WALS feature IDs, the number of answer options, and language coverage. The latter includes the number of languages with at least one valid answer -- i.e., at least one relevant word-order pattern observed in the corpus. While we release the evaluation data for all languages, we perform the evaluation in our paper only on the languages with reasonable statistical support. We determine the threshold of $30$ valid answers as a reasonable trade-off between evaluation on a high number of languages and evaluation with sufficient data support. For feature 144A, the threshold was lowered to $10$, reflecting its generally lower frequency across treebanks (marked with an asterisk in Table  \ref{tab:all-ling-questions-compressed}). 

\subsection{Evaluation Metrics}\label{sec:evaluation-metrics}

During the evaluation phase, the final answers provided by the manually-written Python scripts are treated as the gold standard against which the answers returned by the grammatical analysis system are compared.
We adopt a three-dimensional evaluation procedure covering different aspects of system performance: accuracy of the dominant answer, answer coverage completeness, and distributional fidelity. 
\begin{itemize}
    \item To measure the \textbf{accuracy of the dominant order} prediction, we use the classification accuracy, defined as the ratio between the number of languages with dominant order prediction matching the ground truth, and the total number of evaluated languages. 
    \item To measure the \textbf{order coverage completeness}, we use the F$_1$ score averaged across the evaluated languages. In the expressions below, given the generated distribution and the correct distribution, the options $O_x$ are the outcomes with relative frequency above zero. Precision, recall, and the coverage F$_1$ are defined as follows, where $L$ is the set of evaluated languages.
    \begin{equation}
        Coverage \ F_1 = \frac{1}{|L]}\sum_{L_i \in L} \frac{2 \cdot P_{L_i}  \cdot R_{L_i}}{P_{L_i} + R_{L_i}}
    \end{equation}
    \begin{equation}
        P_{L_i} = \frac{|O_{predicted} \cap O_{correct}|}{|O_{predicted}|}
    \end{equation}
    \begin{equation}
        R_{L_i} = \frac{|O_{predicted} \cap O_{correct}|}{|O_{correct}|}
    \end{equation}
    \item To measure the \textbf{fidelity of the predicted order distribution}, we use the Hellinger distance \citep{Hellinger1909}. The distance is bound between $0$ and $1$, with a lower score indicating better fidelity, i.e. the returned distribution being closer to the manually computed one. The Hellinger distance between two distributions is defined with the following equation, where $P$ is the predicted option distribution and $C$ is the correct option distribution.
    \begin{equation}
        H(P, C) = \frac{1}{\sqrt{2}} \sqrt{\sum_i(\sqrt{P_i} - \sqrt{C_i})^2}
    \end{equation}
\end{itemize}

\section{Agentic LLM System}\label{sec:system}

\begin{figure*}[htbp]
    \centering
    \includegraphics[width=\linewidth]{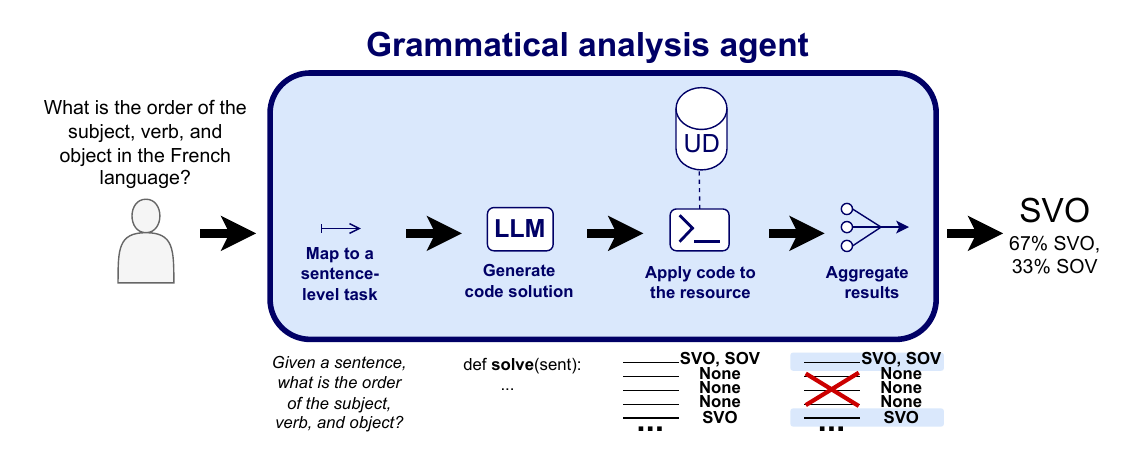}
    \caption{The proposed UDagent system implementation, along with a system trace showing how the system would handle the problem of determining the order of the subject, verb, and object in the French language.}
    \label{fig:agent-implementation}
\end{figure*}

In this section, we describe our grammatical analysis agent prototype. An agent may include various components (e.g., planning, code generation, tool calling), helping it adapt to different needs when dealing with different tasks. To test the feasibility of a grammatical analysis agent, we propose a simple architecture to constrain the initial system complexity. We show its conceptual design in Figure \ref{fig:agent-implementation}, and describe its key components next.

\textbf{Planning.} The proposed system uses a fixed plan, converting the initial general task into an example-level subtask, generating code to solve it, applying it to the corpus, and summarizing the results into a concise answer (majority answer) and an extended answer (answer distribution). 

\textbf{Task conversion.} To convert the general task into an example-level task, its description is edited to include that the task is to be done on a single sentence. For example, based on the initial task of determining the subject, verb, and object order in French, an example-level subtask of determining the subject, verb, and object order \textit{in the given sentence} is created.

\textbf{Linguistic processing.} Based on the example-level task description, the task solution code is generated using an LLM. While we will use open source models in our final approach, in our prototype, we use state-of-the-art models to test the feasibility of the approach. In our current experiments, we use the GPT-5 model \citep{gpt5} with high reasoning effort and low answer verbosity, although this model can be exchanged with another model. 
In the prompt, a description of the data format in UD (CONLL-U) is provided along with the instructions to generate a Python function accepting a single sentence as an argument, and returning one or more of the possible valid answers, or None if the question is not answerable\footnote{We will provide the specific prompts used in the Github repository after the review period.}. The valid answer options are included in the prompt due to the specific formulation of WALS tasks to simplify the evaluation procedure.

\textbf{Resource analysis.} The generated code is executed example-by-example on the chosen UD treebank. More concretely, in our experiments, we execute the generated code on the training splits of UD treebanks if they exist, and the test splits otherwise. None of the ground truth outputs are exposed to the system during the process, allowing us to use test sets without the risk of data contamination. After code execution, each example is assigned an answer, or the value None if the task is not solvable for a given example. For example, a sentence cannot be assigned a subject, verb, and object order if it does not contain all three elements.  

\textbf{Result summarization.} To produce the response, a distribution of the generated example-level answers is computed. The answers with the value None are discarded; thus, the code generation module also implicitly acts as a retrieval filter. The distribution represents the detailed answer returned by the system, while the dominant order prediction is produced by returning the most frequently occurring item in the distribution. For example, if there are $67\%$ examples with the SVO order and $33\%$ with the SOV order, the dominant answer is SVO. 


\section{Evaluation and Analysis}\label{sec:evaluation}
In this section, we evaluate the proposed grammatical analysis system (UDagent) using the presented evaluation framework. First, we begin with a quantitative analysis across all word-order tasks (Section \ref{sec:quantitative-evaluation}), then examine performance patterns across languages (Section \ref{sec:eval-per-lang}), and finally analyse failure cases to identify limitations of the approach (Section \ref{sec:qualitative-evaluation}).

In the quantitative analysis, we measure the accuracy of the computed dominant answer, the coverage completeness, and the distributional fidelity using the metrics presented in Section \ref{sec:evaluation-method}. 
We compare UDagent against two baselines:
\begin{itemize}
    \item \textbf{Majority baseline.} As the evaluated tasks may have an imbalanced ground truth distribution, we include the majority baseline to mitigate the chance for a potential misinterpretation of a high accuracy score. For a given language, the majority baseline predicts the dominant answer for the feature according to WALS. We select the dominant answer from WALS instead of the training distribution as the training set is not available for every language. Its performance can be seen as the lower boundary on acceptable performance.
    \item \textbf{Off-the-shelf GPT-5 baseline.} To test the existing linguistic knowledge in the latest LLMs, and the effect of (not) including external data into a grammatical analysis system, we include the unmodified GPT-5 model \citep{gpt5} as a baseline. We use the model with the reasoning effort set to ``high'' and the answer verbosity set to ``low''.
\end{itemize}

\subsection{Overall Performance}\label{sec:quantitative-evaluation}

\begin{table*}[htbp]
\centering
\begin{tabular}{l c@{\hspace{5pt}}c@{\hspace{5pt}}c@{\hspace{20pt}}c@{\hspace{5pt}}c@{\hspace{5pt}}c@{\hspace{20pt}}c@{\hspace{5pt}}c@{\hspace{5pt}}c}
\toprule
& \multicolumn{3}{c}{(Accuracy $\uparrow$)} & \multicolumn{3}{c}{(Coverage F$_1$ $\uparrow$)} & \multicolumn{3}{c}{(Hellinger dist. $\downarrow$)}\\
Feature & Majority & GPT-5 & UDagent & Majority & GPT-5 & UDagent & Majority & GPT-5 & UDagent \\
\midrule
81A & 
$0.326$ & $0.772$ & $\textbf{0.935}$ &
$0.301$ & $0.783$ & $\textbf{0.826}$ &
$0.703$ & $0.242$ & $\textbf{0.179}$ \\

82A & 
$0.905$ & $0.918$ & $\textbf{0.986}$ &
$0.687$ & $0.918$ & $\textbf{0.968}$ &
$0.303$ & $0.147$ & $\textbf{0.075}$ \\

83A & 
$0.417$ & $0.865$ & $\textbf{0.955}$ &
$0.660$ & $0.870$ & $\textbf{0.955}$ &
$0.560$ & $0.151$ & $\textbf{0.097}$ \\

84A & 
$0.525$ & $0.703$ & $\textbf{0.822}$ &
$0.302$ & $0.734$ & $\textbf{0.908}$ &
$0.644$ & $0.349$ & $\textbf{0.169}$ \\

85A & 
$0.346$ & $0.932$ & $\textbf{0.985}$ &
$0.531$ & $0.850$ & $\textbf{0.924}$ &
$0.657$ & $0.105$ & $\textbf{0.043}$ \\

86A & 
$0.530$ & $\textbf{0.970}$ & $0.955$ &
$0.636$ & $0.864$ & $\textbf{0.904}$ &
$0.466$ & $0.101$ & $\textbf{0.061}$ \\

87A & 
$0.333$ & $0.911$ & $\textbf{0.976}$ & 
$0.572$ & $0.862$ & $\textbf{0.962}$ &
$0.686$ & $0.136$ & $\textbf{0.039}$ \\

88A & 
$0.838$ & $0.878$ & $\textbf{1.000}$ &
$0.815$ & $0.874$ & $\textbf{0.937}$ &
$0.176$ & $0.131$ & $\textbf{0.020}$ \\

89A & 
$0.069$ & $\textbf{0.931}$ & $\textbf{0.931}$ &
$0.457$ & $0.756$ & $\textbf{0.847}$ &
$0.836$ & $0.189$ & $\textbf{0.125}$ \\

90A & 
$0.721$ & $\textbf{0.904}$ & $0.798$ &
$0.744$ & $0.795$ & $\textbf{0.904}$ &
$0.296$ & $\textbf{0.150}$ & $0.182$ \\

94A & 
$0.871$ & $0.903$ & $\textbf{0.936}$ &
$0.634$ & $0.664$ & $\textbf{0.948}$ &
$0.252$ & $0.224$ & $\textbf{0.088}$ \\

144A & 
$0.278$ & $\textbf{0.667}$ & $0.056$ &
$0.180$ & $\textbf{0.559}$ & $0.028$ &
$0.801$ & $\textbf{0.419}$ & $0.971$ \\

144B & 
$\textbf{0.774}$ & $0.642$ & $0.000$ &
$0.335$ & $\textbf{0.569}$ & $0.000$ &
$0.502$ & $\textbf{0.408}$ & $1.000$ \\
\bottomrule
\end{tabular}
\caption{Dominant order prediction accuracy, order coverage completeness ($F_1$), and distributional fidelity (Hellinger distance) scores achieved by the proposed system (UDagent) and the two baselines. The best scores for each task are shown in bold.}
\label{tab:results-all-in-one}
\end{table*}

We present the automated metric scores for our proposed system and the two baselines in Table \ref{tab:results-all-in-one}.

Focusing on the baseline scores in isolation, we can observe that the majority baseline scores vary significantly, showing the different degrees of imbalance across tasks. For example, the accuracy of the majority baseline is $0.838$ for feature 88A, indicating one order of the demonstrative and noun is dominant for most evaluated languages.
The GPT-5 baseline achieves fairly high metric scores across most tasks, achieving the best accuracy score on four, the best coverage F$_1$ on two, and the best Hellinger distance on three tasks.

In general, we can observe that the proposed system achieves a performance improvement against the baselines in the majority of the tasks. More concretely, it achieves the best accuracy scores in $9$, the best coverage F$_1$ scores in $10$, and the best distributional fidelity in $9$ out of $13$ tasks. In these cases, the achieved accuracy scores and the coverage F$_1$ scores are very high, in most cases above $0.9$. Moreover, the Hellinger distances are low (often below $0.1$), indicating the computed distribution closely follows the ground truth distribution.
In two out of $13$ tasks our system fails, and achieves performance worse than the GPT-5 baseline as well as the majority baseline. This can be attributed to the failure of the code generation module which produced code that was syntactically correct but was completely unsuccessful at solving the example-level task (see Section \ref{sec:qualitative-evaluation} for details). 

\subsection{Cross-Linguistic Analysis}\label{sec:eval-per-lang}

To get a better insight into the performance of UDagent, we analyze its behavior by language, more specifically, its accuracy and coverage. As the system fails universally on features 144A and 144B, we exclude them from analysis, focusing on nontrivial performance instead. To estimate the accuracy for a language, we count the number of correctly solved tasks out of tasks for which a language has the ground truth label assigned.

We observe that our system achieves perfect dominant order accuracy for $112$ out of $159$ languages; $37$ achieve perfect coverage of attested orders in addition to perfect accuracy. The former set involves a diverse selection of languages, hinting towards the strong language independence of the produced code. The latter set mostly involves languages with fewer annotated features, making perfect performance easier to achieve. A notable exception is Czech, with perfect accuracy and coverage for all $11$ analyzed features.

\begin{figure*}
    \centering
    \includegraphics[width=\linewidth]{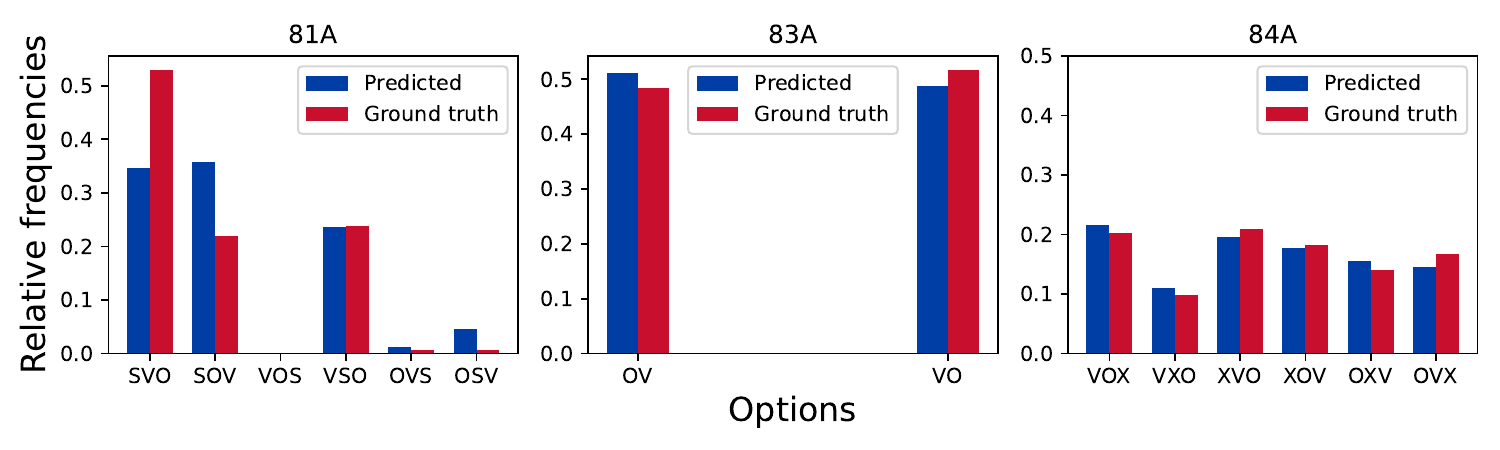}
    \caption{Side-by-side visualization of the predicted and the ground truth intermediate answer distribution for features 81A, 83A, and 84A for Dutch. On features 83A and 84A minor distribution miscalculations result in the wrong dominant answer prediction.}
    \label{fig:distributions-dutch}
\end{figure*}

At the other end of the spectrum, we observe that the system does not perform terribly even on the worst-ranked languages. For two of the five worst-ranked languages with respect to dominant order prediction accuracy (Sinhala, Yupik), very few tasks with ground truth labels are available (2 and 3, respectively). 
For the remaining three, the system achieves accuracy of $0.556$ and greater:
\begin{itemize}
    \item In the case of \textbf{Buryat} and \textbf{Komi Zyrian}, the incorrect answers can be attributed to the very limited size of the UD training splits for these treebanks, which contain few or no instances of the relevant grammatical phenomena.\footnote{The small training set size results from a specific UD policy prioritizing the size of the test set when the treebank size is small.} 
    \item In case of \textbf{Dutch}, the three incorrect dominant order predictions out of nine stem from imprecise distribution estimates. For features 83A and 84A, the dominant options occur with nearly equal frequency, making the outcome highly sensitive to small distributional shifts. In both cases, UDagent slightly reversed their frequency order (see Figure \ref{fig:distributions-dutch}).  
    For feature 81A, the discrepancy reflects a different linguistic interpretation of the  task (see also Section \ref{sec:qualitative-evaluation}): our WALS-inspired ground truth includes only nominal arguments (i.e., subjects and objects expressed as nouns), whereas the agent also counted other parts of speech, such as pronouns -- a difference that is not necessarily erroneous given the simple, WALS-agnostic phrasing of the question given in the prompt.

\end{itemize} 

\subsection{Error and Failure Analysis}\label{sec:qualitative-evaluation}

To better understand the causes of performance variation, we conducted a qualitative inspection of the code generated by the system for the five tasks where it performed below the GPT-5 or majority baseline (86A, 89A, 90A, 144A, and 144B). Examining the intermediate scripts provided insight into how the agent interprets linguistic instructions and where the process fails. Two main types of issues emerged: (1) technical errors that prevented the intended operation of the script, and (2) conceptual mismatches in the linguistic interpretation of the task.

The first category includes logic or data-handling errors that did not break Python syntax but produced incomplete or misleading outputs. For instance, in feature 144A (\textit{Order of subject, object, verb, and negative
word}), there is an error in the generated code when attempting to read the \textit{Polarity=Neg} UD feature from tokens in the treebank. Consequently, no negative words were identified in any treebank and thus the code could not produce appropriate results. Such local flaws had cascading effects, rendering otherwise sound reasoning ineffective.

In contrast, the second category reflects differences in how the model operationalized grammatical concepts given in the prompt. For example, in feature 86A (\textit{What is the order of genitive and noun?}), both our reference implementation and the agent-generated code identified nominal phrases in which the modifying noun bore the \textit{Case=Gen} feature as potential genitive constructions (among other structures). However, while our ground-truth implementation only considered genitive-like constructions labeled with the \textit{nmod} or \textit{nmod:poss} relations (e.g., \textit{the president’s car}), the system also included instances of the \textit{det} relation, which some treebanks use to mark adjectival pronominals expressing possession (e.g. \textit{my car}). As with the Dutch example above, this is not necessarily an error but rather a broader interpretation of the grammatical phenomenon implied by the prompt.



\vspace{1em}

Overall, these analyses highlight two complementary limitations of the current system: the fragility of automatic code generation for complex data operations, and the inherent ambiguity of translating underspecified linguistic definitions into executable form. Both point to the need for future work on incorporating safeguards and clearer linguistic specifications into the prompting and verification stages of grammatical analysis agents. At the same time, the agent’s capacity to devise novel computational strategies for solving linguistic problems also reveals an unprecedented opportunity to move beyond hypothesis-driven corpus analysis, opening new avenues for data-driven discovery.

\section{Conclusion}
\label{sec:conclusion}

This paper presents a framework for corpus-grounded grammatical analysis using agentic LLMs. The proposed system, UDagent, integrates natural-language interpretation, code generation, and corpus-based reasoning, allowing grammatical questions to be answered directly from annotated data. Tested on multilingual word-order variation in UD corpora, the approach proved both scalable and effective, handling over 170 languages with interpretable outputs that closely reflected corpus distributions.

The evaluation showed that even with a relatively simple architecture, the agent achieved strong accuracy and coverage across most tasks, confirming the feasibility of LLM-assisted grammatical research. However, the study has important limitations. The evaluation relied on a single proprietary model (GPT-5), covered only a handful of elementary linguistic phenomena with uneven language coverage, and revealed code-generation failures on complex tasks. Future work should test open and smaller models, expand to multi-step reasoning tasks requiring diverse aggregation strategies, and generalize the architecture -- for instance, by incorporating parsers as tools to process raw text directly.

Despite these limitations, the results demonstrate that LLM-based systems can meaningfully bridge linguistic questions and empirical corpus evidence. By making large-scale grammatical analysis more accessible and interpretable, such systems complement traditional linguistic resources and offer new pathways for cross-linguistic discovery, including the enrichment of typological databases such as WALS with usage-based evidence. As agentic architectures and tool-use capabilities continue to advance, the integration of LLM reasoning with structured linguistic data holds significant promise for transforming how we explore and document grammatical knowledge across the world's languages.





\section*{Acknowledgments}
The work was primarily supported by the Slovene Research and Innovation Agency (ARIS) project GC-0002 and the core research programme P6-0411. The work was also supported by EU through ERA Chair grant no. 101186647 (AI4DH). We acknowledge the support of the EC/EuroHPC JU and the Slovenian Ministry of HESI via the project SLAIF (grant number 101254461). 

\clearpage
\section*{Bibliographical References}\label{sec:reference}

\bibliographystyle{lrec2026-natbib}
\bibliography{lrec2026-example}

\begin{thebibliography}{2}
\expandafter\ifx\csname natexlab\endcsname\relax\def\natexlab#1{#1}\fi

\bibitem[{Dryer and Haspelmath(2013)}]{wals}
Matthew S. Dryer and Martin Haspelmath. 2013.
\newblock \href {https://doi.org/10.5281/zenodo.13950591} {\emph{WALS Online (v2020.4)}}.
\newblock Zenodo.

\bibitem[{Zeman et~al.(2025)}]{11234/1-5901}
Daniel Zeman et~al. 2025.
\newblock \href {http://hdl.handle.net/11234/1-5901} {Universal dependencies 2.16}.
\newblock {LINDAT}/{CLARIAH}-{CZ} digital library at the Institute of Formal and Applied Linguistics ({{\'U}FAL}).

\end{thebibliography}


\begin{thebibliography}{34}
\expandafter\ifx\csname natexlab\endcsname\relax\def\natexlab#1{#1}\fi

\bibitem[{Baylor et~al.(2024)Baylor, Ploeger, and Bjerva}]{Baylor2024}
Emi Baylor, Esther Ploeger, and Johannes Bjerva. 2024.
\newblock \href {https://aclanthology.org/2024.eacl-short.6/} {Multilingual gradient word-order typology from {U}niversal {D}ependencies}.
\newblock In \emph{Proceedings of the 18th Conference of the European Chapter of the Association for Computational Linguistics (Volume 2: Short Papers)}, pages 42--49.

\bibitem[{Beguš et~al.(2025)Beguš, Dabkowski, and Rhodes}]{Begus_2025}
Gašper Beguš, Maksymilian Dabkowski, and Ryan Rhodes. 2025.
\newblock \href {https://doi.org/10.1109/tai.2025.3575745} {Large linguistic models: {Investigating LLMs’} metalinguistic abilities}.
\newblock \emph{IEEE Transactions on Artificial Intelligence}, page 1–15.

\bibitem[{Berdicevskis et~al.(2018)Berdicevskis, Çöltekin, Ehret, von Prince, Ross, Thompson, Yan, Demberg, Lupyan, Rama, and Bentz}]{osti_10119528}
Aleksandrs Berdicevskis, Çağrı Çöltekin, Katharina Ehret, Kilu von Prince, Daniel Ross, Bill Thompson, Chunxiao Yan, Vera Demberg, Gary Lupyan, Taraka Rama, and Christian Bentz. 2018.
\newblock \href {https://doi.org/10.18653/v1/W18-6002} {Using universal dependencies in cross-linguistic complexity research}.
\newblock \emph{Proceedings of the Second Workshop on Universal Dependencies (UDW 2018)}.

\bibitem[{Biber(2009)}]{10.1093/oxfordhb/9780199544004.013.0008}
Douglas Biber. 2009.
\newblock \href {https://doi.org/10.1093/oxfordhb/9780199544004.013.0008} {Corpus-based and corpus-driven analyses of language variation and use}.
\newblock In \emph{The Oxford Handbook of Linguistic Analysis}. Oxford University Press.

\bibitem[{Biber et~al.(2021)Biber, Johansson, Leech, Conrad, and Finegan}]{biber2021grammar}
Douglas Biber, Stig Johansson, Geoffrey~N. Leech, Susan Conrad, and Edward Finegan. 2021.
\newblock \href {https://doi.org/10.1075/z.232} {\emph{Grammar of Spoken and Written English}}.
\newblock John Benjamins Publishing Company.

\bibitem[{Cheung and Crosthwaite(2025)}]{Cheung20052025}
Lisa Cheung and Peter Crosthwaite. 2025.
\newblock \href {https://doi.org/10.1080/09588221.2025.2506480} {{CorpusChat}: integrating corpus linguistics and generative {AI} for academic writing development}.
\newblock \emph{Computer Assisted Language Learning}, 0(0):1--27.

\bibitem[{Crosthwaite and Baisa(2023)}]{CROSTHWAITE2023100066}
Peter Crosthwaite and Vit Baisa. 2023.
\newblock \href {https://doi.org/https://doi.org/10.1016/j.acorp.2023.100066} {Generative {AI} and the end of corpus-assisted data-driven learning? {Not} so fast!}
\newblock \emph{Applied Corpus Linguistics}, 3(3):100066.

\bibitem[{Davies(2025)}]{davies2025aillms}
Mark Davies. 2025.
\newblock \href {https://www.english-corpora.org/ai-llms/english-corpora-with-ai-llms.pdf} {{AI}/{LLM} integration with the corpora from {English-Corpora.org}}.
\newblock Technical report, English-Corpora.org.
\newblock Last edited 25 August 2025.

\bibitem[{de~Marneffe et~al.(2021)de~Marneffe, Manning, Nivre, and Zeman}]{10.1162/coli_a_00402}
Marie-Catherine de~Marneffe, Christopher~D. Manning, Joakim Nivre, and Daniel Zeman. 2021.
\newblock \href {https://doi.org/10.1162/coli_a_00402} {Universal dependencies}.
\newblock \emph{Computational Linguistics}, 47(2):255--308.

\bibitem[{Futrell et~al.(2015)Futrell, Mahowald, and Gibson}]{doi:10.1073/pnas.1502134112}
Richard Futrell, Kyle Mahowald, and Edward Gibson. 2015.
\newblock \href {https://doi.org/10.1073/pnas.1502134112} {Large-scale evidence of dependency length minimization in 37 languages}.
\newblock \emph{Proceedings of the National Academy of Sciences}, 112(33):10336--10341.

\bibitem[{Gerdes et~al.(2021)Gerdes, Kahane, and Chen}]{gerdes:hal-04067297}
Kim Gerdes, Sylvain Kahane, and Xinying Chen. 2021.
\newblock \href {https://doi.org/10.5334/gjgl.764} {{Typometrics: From Implicational to Quantitative Universals in Word Order Typology}}.
\newblock \emph{{Glossa: a journal of general linguistics (2021-...)}}, 6(1):17.

\bibitem[{Gries(2009)}]{Gries2009}
Stefan~Th. Gries. 2009.
\newblock \href {https://doi.org/10.4324/9780203880920} {\emph{Quantitative Corpus Linguistics with R: A Practical Introduction}}, 1 edition.
\newblock Routledge, New York.

\bibitem[{Guo et~al.(2024)Guo, Deng, Wen, Chen, Chang, and Wang}]{ds-agent}
Siyuan Guo, Cheng Deng, Ying Wen, Hechang Chen, Yi~Chang, and Jun Wang. 2024.
\newblock {DS-agent: Automated} data science by empowering large language models with case-based reasoning.
\newblock In \emph{Proceedings of the 41st International Conference on Machine Learning}.

\bibitem[{Hellinger(1909)}]{Hellinger1909}
E.~Hellinger. 1909.
\newblock Neue begründung der theorie quadratischer formen von unendlichvielen veränderlichen.
\newblock \emph{Journal für die reine und angewandte {M}athematik}, 136:210--271.

\bibitem[{Herrera et~al.(2024)Herrera, Corro, and Kahane}]{herrera-etal-2024-sparse}
Santiago Herrera, Caio Corro, and Sylvain Kahane. 2024.
\newblock \href {https://aclanthology.org/2024.lrec-main.1314/} {Sparse logistic regression with high-order features for automatic grammar rule extraction from treebanks}.
\newblock In \emph{Proceedings of the 2024 Joint International Conference on Computational Linguistics, Language Resources and Evaluation (LREC-COLING 2024)}, pages 15114--15125.

\bibitem[{Jumelet et~al.(2025)Jumelet, Weissweiler, Nivre, and Bisazza}]{jumelet2025multiblimp10massivelymultilingual}
Jaap Jumelet, Leonie Weissweiler, Joakim Nivre, and Arianna Bisazza. 2025.
\newblock \href {http://arxiv.org/abs/2504.02768} {{MultiBLiMP 1.0: A} massively multilingual benchmark of linguistic minimal pairs}.

\bibitem[{Kaunisto and Schilk(2024)}]{kaunisto2024challenges}
Mark Kaunisto and Martin Schilk, editors. 2024.
\newblock \href {https://doi.org/10.1075/scl.118} {\emph{Challenges in Corpus Linguistics}}, volume 118 of \emph{Studies in Corpus Linguistics}.
\newblock John Benjamins Publishing Company.

\bibitem[{Leech(1992)}]{leech1992corpora}
Geoffrey Leech. 1992.
\newblock Corpora and theories of linguistic performance.
\newblock \emph{Directions in corpus linguistics}, 1992:105--122.

\bibitem[{Levshina(2019)}]{Levshina+2019+533+572}
Natalia Levshina. 2019.
\newblock \href {https://doi.org/doi:10.1515/lingty-2019-0025} {Token-based typology and word order entropy: A study based on universal dependencies}.
\newblock \emph{Linguistic Typology}, 23(3):533--572.

\bibitem[{Levshina et~al.(2023)Levshina, Namboodiripad, Allassonnière-Tang, Kramer, Talamo, Verkerk, Wilmoth, Rodriguez, Gupton, Kidd, Liu, Naccarato, Nordlinger, Panova, and Stoynova}]{Levshina2023}
Natalia Levshina, Savithry Namboodiripad, Marc Allassonnière-Tang, Mathew Kramer, Luigi Talamo, Annemarie Verkerk, Sasha Wilmoth, Gabriela~Garrido Rodriguez, Timothy~Michael Gupton, Evan Kidd, Zoey Liu, Chiara Naccarato, Rachel Nordlinger, Anastasia Panova, and Natalia Stoynova. 2023.
\newblock \href {https://doi.org/10.1515/ling-2021-0098} {Why we need a gradient approach to word order}.
\newblock \emph{Linguistics}, 61(4):825--883.

\bibitem[{Lewis et~al.(2020)Lewis, Perez, Piktus, Petroni, Karpukhin, Goyal, K\"{u}ttler, Lewis, Yih, Rockt\"{a}schel, Riedel, and Kiela}]{rag}
Patrick Lewis, Ethan Perez, Aleksandra Piktus, Fabio Petroni, Vladimir Karpukhin, Naman Goyal, Heinrich K\"{u}ttler, Mike Lewis, Wen-tau Yih, Tim Rockt\"{a}schel, Sebastian Riedel, and Douwe Kiela. 2020.
\newblock Retrieval-augmented generation for knowledge-intensive {NLP} tasks.
\newblock In \emph{Proceedings of the 34th International Conference on Neural Information Processing Systems}, NIPS '20.

\bibitem[{Liu et~al.(2025)Liu, Chai, Zhu, Tang, Ye, Zhang, Bai, and Chen}]{ml-agent}
Zexi Liu, Jingyi Chai, Xinyu Zhu, Shuo Tang, Rui Ye, Bo~Zhang, Lei Bai, and Siheng Chen. 2025.
\newblock \href {http://arxiv.org/abs/2505.23723} {{ML-Agent: Reinforcing LLM} agents for autonomous machine learning engineering}.

\bibitem[{Lu et~al.(2024)Lu, An, Wang, yang, Kong, Liu, Wang, Lin, Fang, Huang, and Yang}]{lu2024textcqlbridgingnatural}
Luming Lu, Jiyuan An, Yujie Wang, Liner yang, Cunliang Kong, Zhenghao Liu, Shuo Wang, Haozhe Lin, Mingwei Fang, Yaping Huang, and Erhong Yang. 2024.
\newblock \href {http://arxiv.org/abs/2402.13740} {From text to {CQL}: Bridging natural language and corpus search engine}.

\bibitem[{McEnery and Hardie(2011)}]{McEnery_Hardie_2011}
Tony McEnery and Andrew Hardie. 2011.
\newblock \emph{Corpus Linguistics: Method, Theory and Practice}.
\newblock Cambridge Textbooks in Linguistics. Cambridge University Press.

\bibitem[{Nivre et~al.(2020)Nivre, de~Marneffe, Ginter, Haji{\v{c}}, Manning, Pyysalo, Schuster, Tyers, and Zeman}]{UD-2.16}
Joakim Nivre, Marie-Catherine de~Marneffe, Filip Ginter, Jan Haji{\v{c}}, Christopher~D. Manning, Sampo Pyysalo, Sebastian Schuster, Francis Tyers, and Daniel Zeman. 2020.
\newblock {U}niversal {D}ependencies v2: An evergrowing multilingual treebank collection.
\newblock In \emph{Proceedings of the Twelfth Language Resources and Evaluation Conference}, pages 4034--4043.

\bibitem[{{OpenAI}(2025)}]{gpt5}
{OpenAI}. 2025.
\newblock \href {https://cdn.openai.com/gpt-5-system-card.pdf} {{GPT-5} system card}.
\newblock Technical report, OpenAI.
\newblock Accessed: 2025-10-22.

\bibitem[{Sinclair(1991)}]{Sinclair1991CorpusCC}
John Sinclair. 1991.
\newblock \href {https://api.semanticscholar.org/CorpusID:62721047} {Corpus, concordance, collocation}.

\bibitem[{Singh et~al.(2023)Singh, Morris, Aneja, Rush, and Gao}]{singh-etal-2023-explaining}
Chandan Singh, John~X. Morris, Jyoti Aneja, Alexander Rush, and Jianfeng Gao. 2023.
\newblock \href {https://doi.org/10.18653/v1/2023.blackboxnlp-1.3} {Explaining data patterns in natural language with language models}.
\newblock In \emph{Proceedings of the 6th BlackboxNLP Workshop: Analyzing and Interpreting Neural Networks for NLP}, pages 31--55.

\bibitem[{Spencer and Kongborrirak(2025)}]{spencer-kongborrirak-2025-llms}
Piyapath~T. Spencer and Nanthipat Kongborrirak. 2025.
\newblock \href {https://aclanthology.org/2025.coling-main.681/} {Can {LLMs} help create grammar?: {Automating} grammar creation for endangered languages with in-context learning}.
\newblock In \emph{Proceedings of the 31st International Conference on Computational Linguistics}, pages 10214--10227.

\bibitem[{Srivastava et~al.(2023)}]{Srivastava2023BeyondTI}
Aarohi Srivastava et~al. 2023.
\newblock \href {https://api.semanticscholar.org/CorpusID:271601672} {Beyond the imitation game: Quantifying and extrapolating the capabilities of language models}.
\newblock \emph{Trans. Mach. Learn. Res.}, 2023.

\bibitem[{Wei et~al.(2022)Wei, Wang, Schuurmans, Bosma, Ichter, Xia, Chi, Le, and Zhou}]{10.5555/3600270.3602070}
Jason Wei, Xuezhi Wang, Dale Schuurmans, Maarten Bosma, Brian Ichter, Fei Xia, Ed~H. Chi, Quoc~V. Le, and Denny Zhou. 2022.
\newblock Chain-of-thought prompting elicits reasoning in large language models.
\newblock In \emph{Proceedings of the 36th International Conference on Neural Information Processing Systems}.

\bibitem[{Xia et~al.(2024)Xia, Xing, Du, Yang, Feng, Xu, Yin, and Xiong}]{xia-etal-2024-fofo}
Congying Xia, Chen Xing, Jiangshu Du, Xinyi Yang, Yihao Feng, Ran Xu, Wenpeng Yin, and Caiming Xiong. 2024.
\newblock \href {https://doi.org/10.18653/v1/2024.acl-long.40} {{FOFO}: A benchmark to evaluate {LLM}s' format-following capability}.
\newblock In \emph{Proceedings of the 62nd Annual Meeting of the Association for Computational Linguistics (Volume 1: Long Papers)}, pages 680--699.

\bibitem[{Yan and Liu(2023)}]{YanLiu2023}
Jianwei Yan and Haitao Liu. 2023.
\newblock \href {https://doi.org/10.1515/lingvan-2021-0001} {Basic word order typology revisited: A crosslinguistic quantitative study based on {UD} and {WALS}}.
\newblock \emph{Linguistics Vanguard}, 9(1):73--85.

\bibitem[{Yao et~al.(2023)Yao, Zhao, Yu, Du, Shafran, Narasimhan, and Cao}]{yao2023react}
Shunyu Yao, Jeffrey Zhao, Dian Yu, Nan Du, Izhak Shafran, Karthik~R Narasimhan, and Yuan Cao. 2023.
\newblock \href {https://openreview.net/forum?id=WE_vluYUL-X} {React: Synergizing reasoning and acting in language models}.
\newblock In \emph{The Eleventh International Conference on Learning Representations}.

\end{thebibliography}

\section*{Language Resource References}
\label{lr:ref}
\bibliographystylelanguageresource{lrec2026-natbib}
\bibliographylanguageresource{languageresource}

\end{document}